\documentclass[lettersize,journal,twoside]{IEEEtran}
\usepackage{amsmath,amsfonts}
\usepackage{algorithmic}
\usepackage{algorithm}
\usepackage{array}
\usepackage[caption=false,font=normalsize,labelfont=sf,textfont=sf]{subfig}
\usepackage{textcomp}
\usepackage{stfloats}
\usepackage{url}
\usepackage{verbatim}
\usepackage{graphicx}
\usepackage{cite}
\usepackage{booktabs}
\usepackage{multirow}
\usepackage{hyperref}
\usepackage{makecell}

\graphicspath{{images/}}

\begin{document}

\title{Self-Supervised Image Super-Resolution Quality Assessment based on Content-Free Multi-Model Oriented Representation Learning}

\author{Kian Majlessi$^{1}$, Amir Masoud Soltani$^{1}$, Mohammad Ebrahim Mahdavi$^{1}$, Aurelien Gourrier$^{2}$, Peyman Adibi$^{1,2}$
\thanks{Corresponding author: Peyman Adibi.}
\thanks{$^{1}$The authors are with the Department of Artificial Intelligence, 
University of Isfahan, Isfahan, Iran.
Emails:
k.majlessi@mehr.ui.ac.ir,
a.m.soltani@eng.ui.ac.ir,
mohammad.mdv@mehr.ui.ac.ir.
}
\thanks{$^{2}$The authors are with the Univ. Grenoble Alpes, CNRS, LIPhy, Grenoble F-38000, France.
Emails:
aurelien.gourrier@univ-grenoble-alpes.fr,
peyman.adibi@univ-grenoble-alpes.fr.}}

\markboth{Preprint Version}%
{Majlessi \MakeLowercase{\textit{et al.}}: $\mathrm{S^3RIQA}$}


\maketitle

\begin{abstract}
Super-resolution (SR) applied to real-world low-resolution (LR) images often results in complex, irregular degradations that stem from the inherent complexity of natural scene acquisition. In contrast to SR artifacts arising from synthetic LR images created under well-defined scenarios, those distortions are highly unpredictable and vary significantly across different real-life contexts. Consequently, assessing the quality of SR images (SR-IQA) obtained from realistic LR, remains a challenging and underexplored problem. In this work, we introduce a no-reference SR-IQA approach tailored for such highly ill-posed realistic settings. The proposed method enables domain-adaptive IQA for real-world SR applications, particularly in data-scarce domains. We hypothesize that degradations in super-resolved images are strongly dependent on the underlying SR algorithms, rather than being solely determined by image content. To this end, we introduce a self-supervised learning (SSL) strategy that first pretrains multiple SR model oriented representations in a pretext stage. Our contrastive learning framework forms positive pairs from images produced by the same SR model and negative pairs from those generated by different methods, independent of image content. The proposed approach $\mathrm{S^3RIQA}$, further incorporates targeted preprocessing to extract complementary quality information and an auxiliary task to better handle the various degradation profiles associated with different SR scaling factors. To this end, we constructed a new dataset, SRMORSS, to support unsupervised pretext training; it includes a wide range of SR algorithms applied to numerous real LR images, which addresses a gap in existing datasets. Experiments on real SR-IQA benchmarks demonstrate that $\mathrm{S^3RIQA}$ consistently outperforms most state-of-the-art relevant metrics.
\end{abstract}

\begin{IEEEkeywords}
Self-Supervised Learning, Super-Resolution, No-Reference Image Quality Assessment
\end{IEEEkeywords}

\section{Introduction}
\label{sec:introduction}

\begin{figure*}[htbp]
	\centerline{\includegraphics[width=\linewidth]{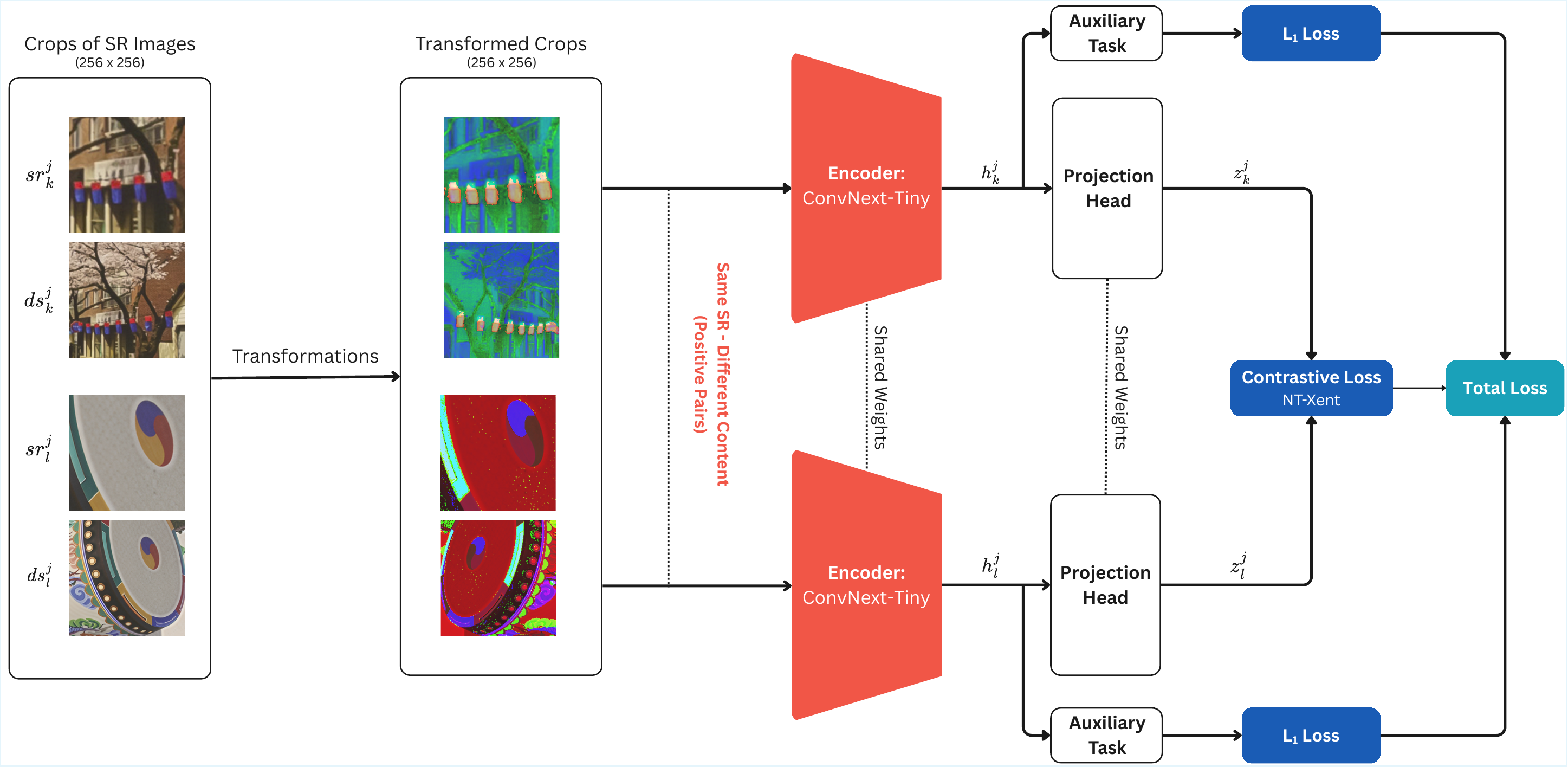}}
    \caption{
    The pretext architecture of $\mathrm{S^3RIQA}$ consists of three components:
    (1) For each SR image $sr_l^j$, a positive sample $sr_k^j$ is generated by taking a random crop from the same SR method but with different content. Both crops are then processed using color–space and random flip transformations.
    (2) Each transformed crop is passed through a symmetric SSL backbone containing an encoder and a projection head, to obtain latent representations $h$ and projected representations $z$, respectively, on which a contrastive loss is computed.
    (3) To encourage learning more robust and scale-aware representations, an auxiliary regression task is applied to the latent features to predict the scaling factor.
    }
    \label{fig:s3riqa}
\end{figure*}

\IEEEPARstart{I}MAGE super-resolution (SR) is a fundamental task in computer vision that aims to enhance the resolution of low-quality images while preserving perceptual quality. Owing to the inherently ill-posed nature of the SR problem, any given low-resolution (LR) input can yield several valid high-resolution (HR) outputs. This ambiguity complicates the establishment of a single standard for image excellence. As a result, SR evaluation frameworks are expected to evaluate SR images based on inherent trade-off between fidelity or mathematical accuracy relative to a ground truth, and the overall perceptual or aesthetic realism of the result \cite{li2025vquala}. Assessing the quality of super-resolved images is therefore difficult due to the intricate degradations involved. Conventional image quality assessment (IQA) techniques which focus mainly on simple degradations such as blur, noise, color distortions, etc. often fail to accurately capture the complex artifacts present in super-resolved images. This is mostly due to complex degradations which commonly result in loss of details, non-uniform blurred edges, and texture inconsistencies \cite{shu2024survey}. Specialized IQA methods for super-resolution (SR-IQA) were therefore designed to accurately evaluate the performance of SR algorithms. Many of those methods compare each super-resolved image with its corresponding HR ground truth, known as the reference, to derive a numerical quality metric. However, obtaining reference images for every LR input is often impractical in real-world applications. Additionally, many methods that rely on pixel-wise comparison with reference images do not necessarily align with human perception of image quality \cite{bianco2018use, lao2025scene}.
As a result, recent research has focused increasingly on developing learning-based blind or no-reference image quality assessment methods (NR-IQA) \cite{shu2024survey,fang2018blind, li2022c, zhang2022joint, zhang2022no, zhang2023boosting, zhou2020blind, shin2024blind, li2024bridging, li2025distilling, li2025few}.
Such methods are mostly based on supervised deep learning and, in principle, do not require HR references within the test/evaluation phase. However, many still require large-scale paired datasets of SR images and quality scores to learn informative features during their training phase. Acquiring such datasets requires changing imaging hardware settings to achieve realistic LR images and collecting statistically meaningful subjective scores, which is often challenging and costly. As a result, most SR-IQA datasets consist of artificially generated LR images, obtained e.g. by down-sampling using blurring filters \cite{ma2017learning}, which facilitates subjective scoring, as demonstrated in Table \ref{tab:sriqa_datasets}. Due to the significant differences between synthetically degraded and real-world LR images, evaluations using synthetic LR images may not accurately reflect the performance of SR algorithms under real conditions \cite{shu2024survey, chen2025real}.
\par
Recently, self-supervised learning (SSL), a subcategory of unsupervised learning, has emerged as a promising approach that improves representation learning in a pretext stage, without requiring paired data, by leveraging the intrinsic structures and statistical properties of images \cite{gui2024survey}. Several efforts have been made to incorporate SSL into NR-IQA \cite{madhusudana2022image, saha2023re, agnolucci2024arniqa} for general low-quality image enhancement. Those approaches typically aim to learn a distortion manifold based on low-quality images. For example, ARNIQA, one of the latest models in this category, employs contrastive learning to bring similar distorted image patches closer together while pushing differently distorted ones apart in the embedding space learned by its encoder \cite{agnolucci2024arniqa}. It constructs its image degradation model by applying random sequences of known distortions—such as brightness changes, compression artifacts, and noise—at varying intensities.\par
However, in real-world blind SR applications \cite{liu2022blind}, including microscopy \cite{jansche2025deep}, remote-sensing \cite{yang2026deep}, astronomical \cite{miao2026frequency}, and underwater \cite{nasir2025self} imaging tasks, the actual degradation sequences that produce low-resolution images are unknown and cannot be accurately modeled simply by combining predefined elementary distortions. Using randomly generated degradation sequences may create distorted images that differ significantly from those produced by existing SR algorithms when applied to real-world LR images. As a result, the latent spaces learned by trained encoders can become misaligned with the actual SR distortions present in the images produced by SR techniques, leading to suboptimal quality metrics and poor performance in SR-specific scenarios.\par
In this work, we introduce the \textit{Self-Supervised Super-Resolution Image Quality Assessment} ($S^3RIQA$), a no-reference IQA method based on SSL designed specifically for blind single-image super-resolution. To the best of our knowledge, this is the first research on the application of SSL techniques in the SR-IQA field. Given that the degradation processes of LR images in real-world are unknown and challenging to model explicitly using conventional transformations, we propose a novel implicit modeling approach, leveraging SSL to learn an SR-specific distortion manifold.

\begin{table*}
    \scriptsize
	\centering
	\caption{Overview of super-resolution image quality assessment datasets.}
	\label{tab:sriqa_datasets}
	\begin{tabular}{lccccccccc}
		\hline
		Dataset & Published & \makecell[l]{Number of \\ reference images} & Scaling factors & \makecell[l]{Number of SR \\ algorithms} & \makecell[l]{Number of \\ SR images} & Synthetic/Realistic & Learning Strategy \\
		\hline
		Yang \emph{et al}. \cite{yang2014single}          & 2014               & 10                              & 2, 3, 4          & 6  & 540   & Synthetic & Supervised \\
		Yeganeh \emph{et al}. \cite{yeganeh2015objective} & 2015               & 13                              & 2, 4, 8          & 8  & 312   & Synthetic & Supervised \\
		CVIU-17 \cite{ma2017learning}                       & 2017               & 30                              & 2, 3, 4, 5, 6, 8 & 9  & 1620  & Synthetic & Supervised \\
		SISRSet \cite{shi2019sisrset}                       & 2019               & 15                              & 2, 3, 4          & 8  & 360   & Synthetic & Supervised \\
		QADS \cite{zhou2019visual}                          & 2019               & 20                              & 2, 3, 4          & 21 & 980   & Synthetic & Supervised \\
		SupER \cite{kohler2019toward}                       & 2019               & 14                              & 2, 3, 4          & 20 & 3024  & Realistic & Supervised \\
		SRIJ \cite{beron2020blind}                          & 2020               & 32                              & 2, 3, 4          & 7  & 608   & Realistic & Supervised \\
		SISAR \cite{zhao2021learning}                       & 2021               & 300                             & 1.5, 2, 2.7      & 6  & 8400  & Synthetic & Supervised \\
		RealSRQ \cite{jiang2022single}                      & 2022               & 60                              & 2, 3, 4          & 10 & 1620  & Realistic & Supervised \\
		SREB \cite{kim2025subjective}                       & 2025               & 30                              & 2, 4             & 7  & 420   & Realistic & Supervised \\
		SRMORSS (Ours)                         & 2026               & 482                             & 2, 3, 4          & 13 & 15374 & Realistic & Unsupervised \\
		\hline
	\end{tabular}
\end{table*}

To achieve this, we treat different SR methods with varying scaling factors as distinct types of degradation, referred to as the \textit{multi-model} \textit{oriented} paradigm. There is also evidence confirming the importance of model awareness for SR-IQA \cite{li2022c}, which supports our proposed model-oriented approach. We then employ SSL to enforce similarity in the representations of outputs generated by the same SR technique with the same scaling factor, even when the input LR images contain different content. In contrast, SSL encourages dissimilar embeddings for outputs produced by different SR techniques or scaling factors, even when the LR images are identical. In this way, the proposed approach implicitly captures the complex and unknown transformations associated with various SR algorithms and scaling factors, which are responsible for both low- and high-frequency artifacts \cite{zhang2022no}. Fig. \ref{fig:s3riqa} provides a schematic representation of the suggested idea.\par
As mentioned before, we focus on the more realistic blind SR-IQA scenario where natural LR images are acquired. There are only a limited number of standard SR-IQA datasets that contain real LR images, as shown in Table \ref{tab:sriqa_datasets}. These datasets also typically contain a limited number of images, making them inadequate for the SSL pretext training phase. Larger datasets are therefore required, which are usually drawn from realistic SR datasets \cite{chen2022real}, such as the well-known and widely used RealSR \cite{cai2019toward}. However, the later obviously doesn't include super-resolved versions of LR images nor their ground-truth quality annotations, since it has not been designed for SR quality assessment. Consequently, while RealSR cannot be employed for supervised downstream tasks, it serves as an appropriate choice for our SSL-based unsupervised pretext task.\par
To overcome those limitations, we introduce a new dataset, referred to as SRMORSS, intended for pretext training of the $\mathrm{S^3RIQA}$ model. This dataset is derived from the content of RealSR by applying several modern and traditional SR techniques to its LR images to obtain SR data. To evaluate generalization and robustness, we also incorporate existing real SR-IQA datasets, which contain subjective quality annotations for supervised downstream training. In this stage, a simple regressor is trained on these datasets to predict the corresponding quality scores. The proposed $\mathrm{S^3RIQA}$ model demonstrates its effectiveness in these SR-IQA datasets that were not used during its representation learning, highlighting the strong domain adaptation capabilities of the model\footnote{The source code, dataset, and material associated with this project can be found at \href{https://github.com/kianmajl/Self-Supervised-Image-Super-Resolution-Quality-Assessment}{https://github.com/kianmajl/Self-Supervised-Image-Super-Resolution-Quality-Assessment}}. Our contributions are summarized as follows.
\begin{enumerate}
    \item We propose the first SSL approach to assess the quality of SR images generated from real LR inputs.
    \item We present a new dataset for SSL-based SR-IQA, which is sufficiently large and diverse, covering a wide range of SR methodologies, to enable reliable pretext training under realistic resolution degradation scenarios.
    \item We introduce a novel contrastive learning framework for SR-IQA, where positive pairs originate from the same SR model independent of image content.
    \item An auxiliary scaling factor prediction task is used to enhance the representations derived from the pretext encoder.
    \item $\mathrm{S^3RIQA}$ outperforms no-reference SR-IQA metrics in existing benchmarks using multiple SR-IQA datasets, with realistic LR images.
\end{enumerate}

The remainder of the paper is organized as follows: Section \ref{sec:related_work} presents a review of the related literature. Section \ref{sec:dataset} describes the proposed dataset. Section \ref{sec:methodology} outlines our methodology, detailing the procedures and techniques employed in this study. Section \ref{sec:results} discusses the experimental results and the analysis of our findings. Finally, Section \ref{sec:conclusion} provides the concluding remarks and suggests directions for future research.

\section{Related Work}
\label{sec:related_work}
Existing SR-IQA methods are broadly classified into full-reference (FR), reduced-reference (RR), and no-reference (NR) approaches, depending on their use of pristine HR images \cite{shu2024survey}. In this section, we briefly review the major categories of SR methodologies to highlight emerging realistic SR scenarios that require robust SR-IQA metrics. We then summarize existing IQA methods that leverage SSL. Next, we briefly review SR-IQA methods, with a primary emphasis on NR approaches, as they are more compatible with the realistic conditions considered in this paper. Finally, we introduce current SR-IQA datasets that contain real LR images along with subjective quality assessment scores.

\subsection{Super-resolution algorithms}

Single image super-resolution (SISR) is a fundamental low-level computer vision problem aimed at recovering an HR image from its LR counterpart. While early methods relied on analytical interpolation techniques, the field has been dominated recently by deep learning approaches, due to their remarkable capacity to learn intricate image priors from data.
Foundational works such as SRCNN \cite{dong2015image} and VDSR \cite{kim2016accurate} pioneered the use of convolutional neural networks (CNNs) with upsampling layers to achieve state-of-the-art performance at the time.\par

The introduction of Generative Adversarial Networks (GANs) further revolutionized the SR field. Real-ESRGAN \cite{wang2021real} enhanced SR for real-world degraded images by extending ESRGAN \cite{wang2018esrgan} with a U-Net discriminator using spectral normalization and a high-order degradation modeling process.\par

Implicit neural representations (INRs) \cite{mildenhall2021nerf} enables arbitrary-scale and even unseen-scale SR within a single model by parameterizing signals continuously in comparison to conventional representations. LIIF \cite{chen2021learning} unfolded a low-resolution feature map into a continuous representation using local implicit image functions. A follow-up work, DIINN \cite{nguyen2023single}, developed a dual interactive implicit network for SISR. HIIF \cite{jiang2025hiif} advanced this paradigm by replacing standard positional encoding with a hierarchical version that captures multi-scale frequencies, while adding a multi-scale decoder and efficient multi-head linear attention to better mix local and global features.\par

Vision Transformers (ViTs) and hybrid CNN-Transformer architectures have recently dominated benchmark leaderboards by capturing long-range dependencies critical for texture restoration. SwinIR \cite{liang2021swinir} introduced a Swin Transformer backbone with residual connections, achieving state-of-the-art results while remaining efficient.\par

Another promising area in SISR is the use of graph neural networks (GNNs). IPG \cite{tian2024image} constructed flexible pixel-wise graphs that adaptively assign higher node degrees to detail-rich regions, thereby breaking the spatial rigidity inherent in CNNs and window-based attention mechanisms. IPG models images using individual pixel nodes and builds local/global graphs to capture both neighborhood and long-range dependencies.\par

As diffusion models \cite{sohl2015deep} became popular, SR methods started to use them as well. InvSR \cite{yue2025arbitrary} reformulated diffusion inversion to leverage the rich image priors encapsulated in large pre-trained diffusion models. Their method uses a Partial noise Prediction (PnP) strategy and a dedicated deep noise predictor, enabling a flexible sampling mechanism that supports an arbitrary number of steps.\par

Given the rising demand for lightweight SR models that utilize less than a million parameters, several efficient approaches emerged.
SeemoRe \cite{zamfir2024see}, an efficient SR model employing Mixture-of-Experts (MoE), mines and integrates expert modules at multiple levels to capture both rank-wise and spatial-wise informative features, enabling more accurate detail reconstruction with reduced computational cost compared to prior heavy architectures. CATANet \cite{liu2025catanet} proposed a lightweight SR method, using a content-aware token aggregation module to aggregate similar tokens together into content-aware regions. In addition, it incorporates an intra-group self- and cross- attention to enable long-range information and enhance global information interaction, respectively.\par

Deep supervision has emerged as an effective strategy to improve network training by alleviating vanishing and exploding gradients and promoting better convergence. In SR, it is commonly combined with skip connections to preserve spatial information across network depths. Aggregating predictions from intermediate supervised layers enables the model to exploit multi-level features, resulting in more accurate and perceptually natural HR outputs. \cite{li2025comprehensive}

\subsection{Super-resolution image quality assessment}
This subsection provides a short review on several recent SR-IQA methods categorized according to their use and reliance on reference images, categorized in the beginning of this section into FR, RR, and NR categories.
Among FR methods, VQAD \cite{zhou2019visual} measures texture, structure, and high-frequency similarities using SIFT descriptors, structure tensors, and high-frequency filtering.
A-FINE \cite{chen2025toward} formulates quality as an adaptive linear combination of fidelity and prior-based terms derived from maximum a posterior (MAP) estimation. RR paradigms are less common \cite{yun2023you, dost2022reduced}.
As an example, PFIQA \cite{lin2024perception} notably extracts deep features from both the SR image and an upsampled LR image using pretrained ViT and ResNet backbones, fusing them separately via an adaptive module.\par
The majority of recent works focus on the more practical NR setting. Early hand-crafted approaches include NRQM/SRIQA \cite{ma2017learning}, which combines local DCT-based generalized Gaussian fits, global steerable-pyramid \cite{simoncelli1992shiftable} Gaussian scale mixtures, and PCA-derived spatial discontinuity features with two-stage random forest regression, and KLTSRQA \cite{jiang2022single}, which applies the Karhunen-Loève transform in opponent color space followed by asymmetric generalized Gaussian distribution (AGGD) modeling \cite{lasmar2009multiscale} and SVMRank \cite{joachims2006training} regression.\par

Deep NR methods are currently the main SR-IQA approaches. SGH \cite{fu2023scale} introduces a scale-conditioned hypernetwork that dynamically adjusts existing IQA models. Zhang \emph{et al}. \cite{zhang2022no} suggested separating high- and low-frequency distortion maps for processing through a twin non-shared ResNet-101 backbone. Hybrid Vision Transformer and Convolutional Neural Network (HVTCNN) \cite{li2025hybrid} integrates hierarchical ViT tokens with CNN refinement. PSCT \cite{zhang2023perception} uniquely combines reference-based and no-reference branches to take advantage of perceptually superior results when references are available.

CN-BSRIQA \cite{rehman2024cn}, a cascaded CNN-DBN network for blind SR-IQA, extracts content-independent low-level features from image patches through enhanced visual quality prediction. $\mathrm{Q_{ODA}}$ \cite{beron2020blind} uses a content-aware quantization framework that dynamically adjusts bit-width allocations across spatial regions and feature channels based on local image content and quantization sensitivity. It thereby balances computational efficiency with high-fidelity image restoration by minimizing quantization error where it would most degrade quality.

C$^2$MT \cite{li2022c} introduces a multi-task transformer that jointly learns SR-image quality prediction and SR-algorithm classification to exploit the mutual information between perceptual quality and algorithm classes. It further incorporates supervised contrastive learning and an active pseudo-label generation strategy to obtain more discriminative and credible quality-aware representations.

Although FR and RR methods generally achieve higher accuracy when references exist, NR approaches remain the only viable option for real-world deployment, where there is no access to HR references, and demonstrate substantially higher generalization to unseen artifacts, along with improved robustness under labeled-data scarcity.

\subsection{Self-supervised learning image quality assessment}
Recent self-supervised IQA methods have eliminated the reliance on human annotations through various strategies. A pioneer work, CONTRIQUE \cite{madhusudana2022image}, proposed an SSL CNN-based NR-IQA method trained on unlabeled images with synthetic and authentic distortions via a contrastive pairwise objective that predicts distortion type and degree as an auxiliary task. Re-IQA \cite{saha2023re} introduced an unsupervised mixture of experts framework that leverages two ResNet-50 encoders for complementary content-aware and quality-aware feature extraction. It has been trained via contrastive learning with MoCo-v2, a new augmentation protocol and an intra-pair image-swapping scheme. ARNIQA \cite{agnolucci2024arniqa} instead adopts the SimCLR contrastive approach on a ResNet-50, constructing a content-agnostic distortion manifold through cross-image patch pairings under identical synthetic multi-degradations, markedly departing from Re-IQA’s intra-image strategy. TRIQA \cite{sureddi2025triqa} introduced a ranked triplet-based strategy paradigm with a margin loss for extracting quality-aware features, achieving the state-of-the-art performance on user-generated content datasets. RankCORE \cite{joshi2025rankcore} proposed a lightweight model trained with self-supervised adaptive ranking (SSAR) in synthetically degraded ordered pairs, highlighting dynamic margins for subtle degradations discrimination.

\subsection{Super-resolution image quality assessment datasets}
\label{sec:sriqad}
As mentioned earlier, we focus on SR-IQA within realistic LR image capture scenarios. Several real-world SR-IQA datasets have introduced authentic degradations by avoiding synthetic bicubic downsampling. The SREB dataset \cite{kim2025subjective} provides 420 SR images from 30 genuine $720\times540$ broadcast clips, enhanced by 7 SR methods and rated via pairwise comparisons by 51 subjects on a single display. The RealSRQ dataset \cite{jiang2022single} contains 180 LR-HR pairs captured with specific cameras at varying focal lengths ($\times2$, $\times3$, $\times4$ real optical zoom), aligned with RealSR registration, from which 1,620 SR outputs of 10 SISR algorithms are scored using Bradley-Terry (B-T) models derived from 65,400 pairwise votes by 60 observers. The SRIJ dataset \cite{beron2020blind} extends the hardware-binned images by cropping them into 32 aligned scenes, producing 608 authentic grayscale SR images at $\times2$, $\times3$ and $\times4$ scales, using 8 SR algorithms, with mean opinion score (MOS) computed from single-stimulus absolute category ratings by 36 valid subjects after Z-score normalization and rescaling to the $[0,100]$ interval.

\section{New SRMORSS Dataset}
\label{sec:dataset}
The reconstructed SR images contain subtle artifacts that differentiate them from genuine HR images. Those artifacts are generally harder to detect, and accurately identifying them is essential to develop more advanced SR-IQA models. Taking into account the variety of methods employed for super-resolving LR images, this study presents a new SR-IQA dataset named \textit{Super-Resolution Model-Oriented Realistic Self-Supervision} {(SRMORSS)} dataset \cite{majlessi_2026_18479156}. Our database contains LR images of real scenes and their SR counterparts generated using a broad distribution of classic and modern SR algorithms. The LR images are sourced from the RealSR dataset and are acquired under realistic conditions. Our dataset contains the largest number of SR images among existing SR-IQA datasets, making it particularly valuable for SSL tasks that benefit from large-scale and diverse image collections.

\subsection{SR methods used for dataset development}
A wide range of objectives, applications and architectural frameworks have been introduced for SISR tasks. Accordingly, the degradations introduced by different SR models are expected to vary, reflecting a model-oriented perspective in which each SR method exhibits unique degradation patterns in its reconstructed images. These differences are more pronounced among SR methods from fundamentally different categories and less evident among methods based on similar techniques. Considering the variety of those degradations and their potential link to the specific SR techniques used, as well as their categories \cite{zhou2022quality}, the $\mathrm{SRMORSS}$ dataset is crafted to include SR methods from various categories: interpolation, CNN, GAN, Transformer, GNN, diffusion, INR, and lightweight approaches of CNN combined with attention. The selected methods are listed in Table \ref{tab:sr_methods}, treating the same SR techniques with different scaling factors as distinct models, since their performance and output quality strongly depend on the chosen scale. In this way, exposing the model to all these variations during the pretext phase of the SSL framework can facilitate learning a rich \textit{SR degradation manifold}, thus simplifying the downstream task of assigning quality scores to new SR-degraded images.

\subsection{Details of SRMORSS dataset}
\begin{figure*}[htbp]
	\centerline{\includegraphics[width=\linewidth]{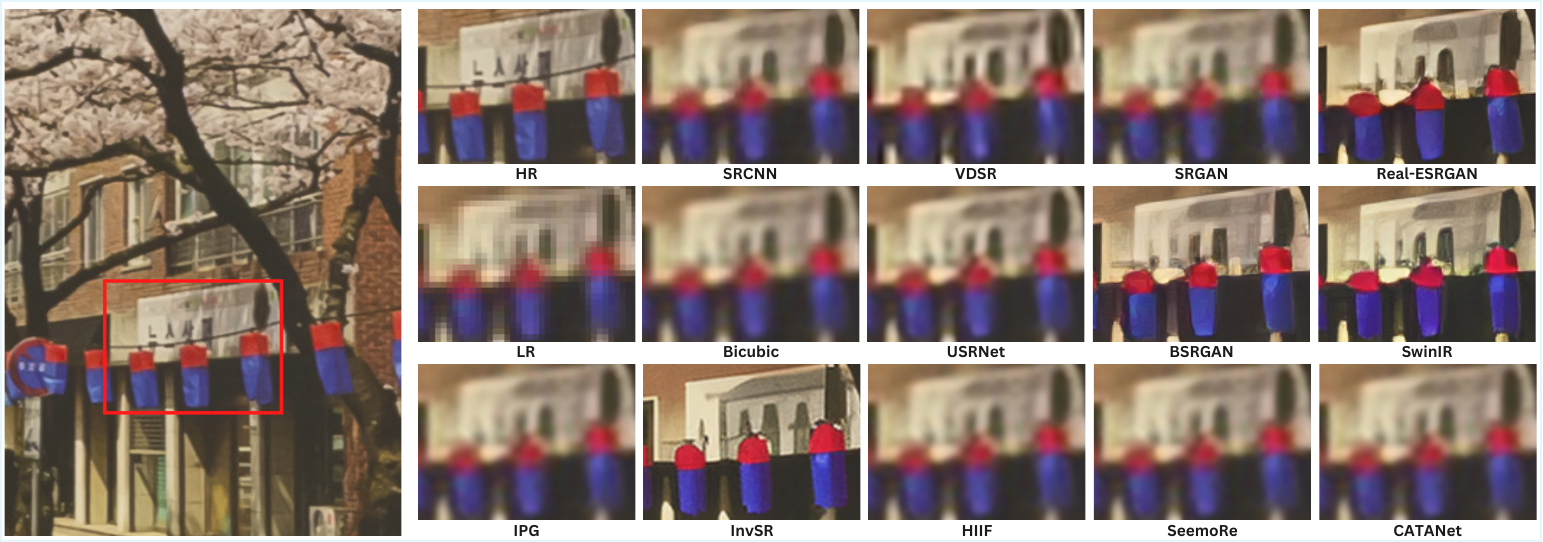}}
	\caption{Visual comparison between the HR image (left) and the corresponding SR images (right) with a scaling factor of 4.}
	\label{fig:s3riqad}
\end{figure*}

The new SRMORSS dataset comprises 1,446 LR images sourced from the RealSR dataset and 15,374 SR reconstructed images generated by 13 various SR methodologies in three upscaling factors: $\times2$, $\times3$ and $\times4$.
This diverse collection and substantial quantity of SR images in SRMORSS yields a robust dataset well-suited to integration into pre-training pipelines of SR-related tasks. In particular, as mentioned above, it is well suited for the pretext task in our SSL-based SR-IQA method. Fig. \ref{fig:s3riqad} displays representative samples from the dataset, highlighting the variety of distortions related to SR.

\begin{table}[t]
    \scriptsize
	\centering
	\caption{Methods used for generating SR images in the SRMORSS dataset.}
	\label{tab:sr_methods}
	\begin{tabular}{lcccc}
		\hline
		Method                           & Type & Year & Scaling Factor \\ 
		\hline
		Bicubic                                   & Interpolation & -             & 2, 3, 4                 \\ \hline
		SRCNN \cite{dong2015image}                 & CNN           & 2015          & 2, 3, 4                 \\
		VDSR \cite{kim2016accurate}                & CNN           & 2016          & 2, 3, 4                 \\
		USRNet \cite{zhang2020deep}                & CNN           & 2020          & 2, 3, 4                 \\ \hline
		SRGAN \cite{ledig2017photo}                & GAN           & 2017          & 2, 4                    \\
		BSRGAN \cite{zhang2021designing}           & GAN           & 2021          & 2, 4                    \\
		Real-ESRGAN \cite{wang2021real}            & GAN           & 2021          & 2, 3, 4                 \\ \hline
		SwinIR\footnotemark \cite{liang2021swinir} & Transformer   & 2021          & 2, 3, 4                 \\ \hline
		IPG \cite{tian2024image}                   & GNN           & 2024          & 4                       \\ \hline
		InvSR \cite{yue2025arbitrary}              & Diffusion     & 2025          & 4                       \\ \hline
		HIIF \cite{jiang2025hiif}                  & INR           & 2025          & 2, 3, 4                 \\
		\hline
		SeemoRe\footnotemark \cite{zamfir2024see}               & \makecell[l]{CNN + MoE \\ Lightweight}   & 2024          & 2, 3, 4                 \\
		CATANet \cite{liu2025catanet}              & \makecell[l]{CNN + Attention \\ Lightweight}   & 2025          & 4                       \\
		\hline
	\end{tabular}
\end{table}

\footnotetext[2]{For SwinIR, scales $\times2$ and $\times4$ use real-world SR weights, while scale $\times3$ relies on classical SR weights because real-world weights are not available for that scale, and we additionally generate SR images using the SwinIR-Large model for the $\times4$ scaling factor.}

\footnotetext{The tiny version of the SeemoRe model is used for generating SR images.}

\section{Methodology}
\label{sec:methodology}

This section presents a comprehensive description of the constituent modules of the proposed $\mathrm{S^3RIQA}$ model and discusses how SR-IQA effectively exploits the SSL framework. 

\subsection{Problem formulation}
$\mathrm{S^3RIQA}$ formulates the IQA task for the super-resolved versions of real LR images using an SSL objective.
For pretext task, our proposed dataset comprises three different image sets:
\begin{equation}
    D=\{lr_i, hr_i, sr_i^j | \forall i \in 1,...,N \;\; \\\\ \text{and} \;\; \forall j \in 1, ..., |\text{SR}|\}
\end{equation}
where $D$ denotes the dataset, $N$ denotes the number of reference images, $|\text{SR}|$ denotes the total number of SR methods employed, $hr_i \in \mathbb{R}^{H\times W\times C}$ denotes the HR counterpart of the LR image $lr_i \in \mathbb{R}^{H/s \times W/s \times C}$ with $H$, $W$, and $C$ being the height, width, and channel number of HR image, respectively, $s$ being the scaling factor of LR image, and
$sr_i^j$ is the output of the $j$-th SR method applied to the $i$-th LR image:
\begin{equation}
	sr_i^j = \text{SR}_j(lr_i) \in \mathbb{R}^{H\times W\times C}
\end{equation}

\subsection{Multiscale crop construction}

Motivated by previous work demonstrating the effectiveness of multiscale information for quality prediction \cite{madhusudana2022image, agnolucci2024arniqa, chen2024perception}, the $\mathrm{S^3RIQA}$ model employs both the original and half-scale (downsampled by a factor of 2 along both dimensions) versions of each image during training and testing. The downsampled version $ds_i^j$ is constructed using the LANCZOS resampling method \cite{burger2010principles} from $sr_i^j$:
\begin{equation}
   ds_i^j = \text{LANCZOS}(sr_i^j) \in \mathbb{R}^{H/2\times W/2\times C} 
\end{equation}

Due to the varying spatial dimensions of the input SR images and the substantial computational resources required to process them in full, we adopt a cropping strategy. During training, a random crop of $256 \times 256$ pixels is sampled from each SR image and its downsampled counterpart at each iteration and used for the pretext task. Although a cropped patch may not perfectly reflect the perceived quality of the SR image, we assume that the underlying image distribution remains effectively unchanged by this cropping operation, during many iterations. As a result, all images in the dataset are included at both scales, effectively doubling the size of the training dataset while providing the model with valuable multiscale property.\par

\subsection{Training}
A conventional contrastive SSL pipeline is built on the comparative analysis of two distinct augmentations derived from the same input sample.
However, the common data augmentations employed in SSL objectives often introduce additional distortions to individual images, thereby degrading crucial information that is essential to achieve strong performance in our downstream IQA task.
To address these limitations and establish meaningful relationships between samples by leveraging the inherent structure of the dataset, the proposed method introduces a \textit{relational pretext task} that depends solely on the SR methodology and the scaling factor used to generate each sample, independent of the image content itself.\par
Each mini-batch $B$ is comprised from the pairs of SR images and their downsampled counterparts. Specifically, to construct positive view pairs for the pretext training given a mini-batch, $\mathrm{S^3RIQA}$ randomly selects two SR images produced by the same technique and scaling factor, but depicting different content, thereby forming a \textit{same-SR-different-content} relation:
\begin{equation}
	(sr_k^j, sr_l^j) \sim D \;\; where \;\; l\ne k 
\end{equation}
where $\sim$ denotes the sampling from the dataset introduced in equation (1) (here from SR images).
These positive pairs force the network to identify subtle distinctions between samples generated by a specific SR method independently of their semantic content, thus focusing on shared degradation patterns inherent to that SR method.\par

Motivated by the CONTRIQUE framework \cite{madhusudana2022image}, a color space conversion transformation is used, randomly transforming the SR image into a new color space
$Color\!\!=\!\!\{\text{Grayscale}, \text{RGB}, \text{LAB}, \text{HSV}, \text{MS}\}$. Note that this transformation does not introduce additional degradation to images.

All remaining images in the same training mini-batch are treated as negative pairs, although some of them may originate from the same SR method, following the same principle as in typical contrastive learning objectives.

\subsection{$S^3RIQA$ architecture}
The core architecture of the $\mathrm{S^3RIQA}$ pretext stage is adapted from the SimCLR self-supervised symmetric learning framework \cite{chen2020simple}, as shown in Fig. \ref{fig:s3riqa}. An encoder along with a projection head process two images of the same SR methodology, but with different contents, with the objective of minimizing their representational distance.
A ConvNeXt-Tiny \cite{liu2022convnet} with its pretrained weights is used as an encoder $f_{\theta}(.)$ to get 768-dimensional latent representations, since it exhibits higher performance compared to typical ResNets:
\begin{align}
    h_k^j = f_{\theta}(sr_k^j) \quad \text{,} \quad
    h_l^j = f_{\theta}(sr_l^j)
\end{align}
Subsequently, the latent representations $h_k^j$ and $h_l^j$ from the $j$th SR method are passed through a two-layer multilayer perceptron (MLP) $g_{\omega}(.)$ that serves as the projection head, which produces 128-dimensional projected representations:
\begin{align}
	z_k^j = g_{\omega}(h_k^j) = W_2GELU(W_1 h_k^j), \notag \\
	z_l^j = g_{\omega}(h_l^j) = W_2GELU(W_1 h_l^j)
\end{align}
where $W_1 \in \mathbb{R}^{512 \times 768}$ and $W_2 \in \mathbb{R}^{128 \times 512}$ are weights of the projection head.
The resulting projected representations $z_k^j$ and $z_l^j$ are utilized in normalized temperature-scaled cross-entropy loss (NT-Xent) \cite{chen2020simple} to minimize the distance between images generated by the same SR method and scaling factor despite their differing content. In addition to this contrastive objective, we incorporate an auxiliary task designed to further guide representation learning, as detailed in the following subsection.

\begin{figure}[htbp]
	\centerline{\includegraphics[width=\linewidth]{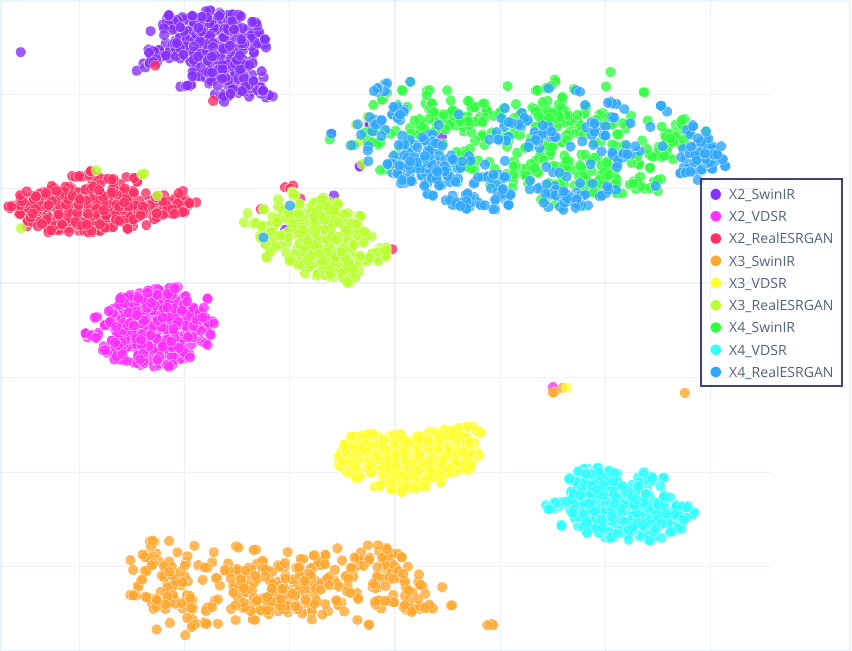}}
	\caption{Partial visualization of the latent space using t-SNE algorithm \cite{maaten2008visualizing} showing the representations of the images of SwinIR, VDSR, RealSRGAN models with scaling factors of $\times2$, $\times3$ and $\times4$. The clusters indicate content-free, model-oriented embeddings that drive the superior performance of $\mathrm{S^3RIQA}$.}
	\label{fig:tsne}
\end{figure}

Fig. \ref{fig:tsne} provides a representation of the latent space of a subset of the SRMOSS dataset, comprising three SR models (SwinIR, VDSR, RealSRGAN) evaluated with three different scaling factors ($\times2$, $\times3$, $\times4$). The proposed contrastive learning approach clearly allows discriminating embeddings corresponding to different SR methods, which is largely independent of image content. Moreover, different scaling factors tend to occupy distinct regions of this SR distortion manifold, due to the auxiliary task described in the next subsection. Together, these properties facilitate downstream quality prediction based on the learned representations. We observe that the embeddings of SwinIR and Real-ESRGAN on the $\times4$ scale are partially intermingled, which is consistent with the visual similarity between these two methods at this scale, as shown in Fig. \ref{fig:s3riqad}, showcasing an additional quality-sensitive advantage of the proposed architecture. Considering all SR models in Table \ref{tab:sr_methods} therefore results in a rich and informative latent space that captures characteristic SR model behaviors independent of image content. This is expected to provide robust representations for any arbitrary image content and given SR models.\par

\subsection{Auxiliary task}
An auxiliary task is implemented using a lightweight MLP to draw out more discriminative representations from the images. This network takes the latent representations generated by the encoder and is trained end-to-end within the pretext task to predict the scaling factor linked to each image as follows:
\begin{equation}
	\hat{s}_i^j = W_4 GELU(W_3 h_i^j)
\end{equation}
where $W_3 \in \mathbb{R}^{512\times 768}$ and $W_4 \in \mathbb{R}^{1\times 512}$ are the learnable weights of the auxiliary task, the GELU function provides the suitable nonlinearity, and $\hat{s}_i^j$ is the predicted scaling factor for the input image corresponding to $h_i^j$.\par
An $\mathrm{L_1}$ loss function is utilized for the auxiliary task to quantify the discrepancy between the predicted scaling factor and the corresponding ground-truth value.
\begin{equation}
    L_{aux} = \frac{1}{|B|}\sum_{s_i^j \hspace{0.5mm} \text{of} \hspace{0.5mm} sr_i^j \in B} |\hat{s}_{i}^j - s_{i}^j|
\end{equation}
where $B$ is the mini-batch and $s_i^j$ denotes the scaling factor of the sample from the $i$-th LR image super-resolved by the $j$-th SR method.

\subsection{Objective function}
The contrastive loss for a mini-batch is computed over positive pairs constructed by selecting an image generated by the same SR method but with distinct content as follows (negative examples are not explicitly sampled):
\begin{equation}
\label{eq:simclr_objective}
	l^j_{l,k} = -log \frac{exp(sim(z_l^j, z_k^j)/\tau)}{\sum_{\substack{ z_m^p \hspace{0.5mm} \text{of} \hspace{0.5mm} sr_m^p \in B \\ z_m^p \ne z_l^j}} exp(sim(z_l^j, z_m^p)/\tau)}
\end{equation}
where $sim$ is a similarity measure (cosine similarity is selected in this work), and $\tau$ is a temperature hyper-parameter. 
Then it is summed over the mini-batch $B$ for all SR images in it:
\begin{equation}
    L_{contr} = \sum_{j,l,k} l^j_{l,k}
\end{equation}
where $j$ and $l$ denote the index of the SR method and the LR image of an SR image in the mini-batch, respectively, and $k$ specifies the content of its positive pair.
The total pretext task loss for the current mini-batch is then computed as the sum of the contrastive loss and auxiliary loss, as follows:
\begin{equation}
	L_{total} = L_{contr} + L_{aux}
\end{equation}

\subsection{Downstream task}

\begin{figure}[htbp]
	\centerline{\includegraphics[width=\linewidth]{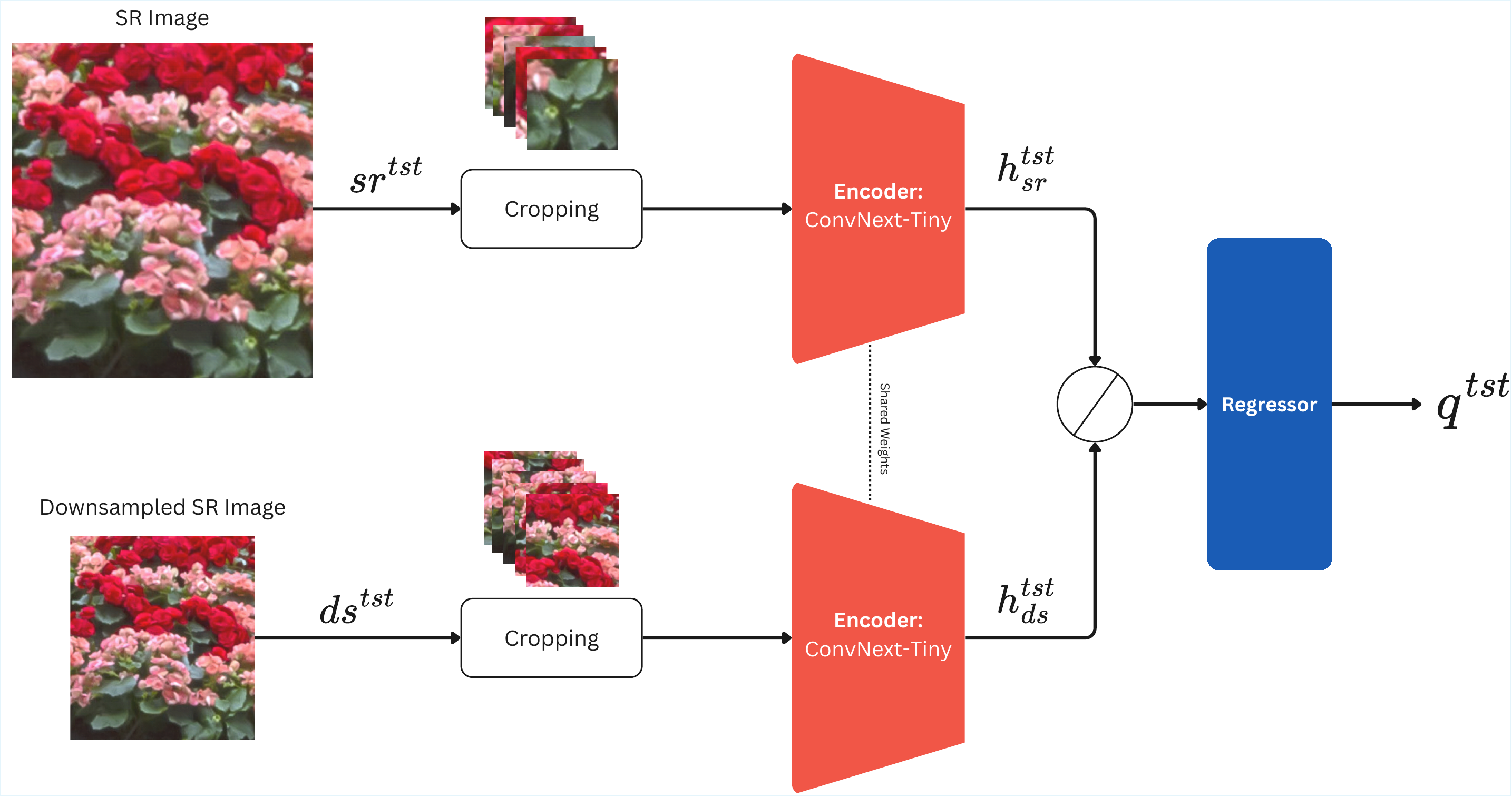}}
	\caption{Evaluation phase of the downstream stage. A test sample $sr^{tst}$ and its downsampled version, first undergoes a cropping step. All crops are then fed into the encoder trained during the pretext stage. The embeddings of each crop and its downsampled counterpart are concatenated (denoted by the symbol $\oslash$) and provided as input to the regression model trained on $D^{trn}$ dataset. The final predicted quality score $q^{tst}$ is computed by averaging the predictions across all crops.}
	\label{fig:s3riqa_downstream}
\end{figure}

A regression model is trained on several realistic supervised SR-IQA datasets (Table \ref{tab:sriqa_datasets}) to predict a quality score for a given test SR image (Fig. \ref{fig:s3riqa_downstream}). Different regression models can be employed in this stage, and several were evaluated, including ridge regression, MLP, and support vector regression (SVR). Interestingly, the simplest model, ridge regression, achieved the best performance in the experiments. This outcome is expected, as the embedding space learned during the pretext stage is relatively simple and exhibits a clear monotonic separation (cf. Fig. \ref{fig:tsne}). Such a structure enables linear models to perform effectively while mitigating overfitting, which may arise when using more complex regression methods.\par

The dataset is split into training and testing sets, as 
$D_{trn}\!\!=\!\!\{(sr_i^{trn},q_i^{trn}) | \forall i \in 1,...,N^{trn}\}$
and $D_{tst}\!\!=\!\!\{(sr_i^{tst},q_i^{tst}) | \forall i \in 1,...,N^{tst}\}$, with $q_i^{trn}$ and $q_i^{tst}$ being the ground-truth quality scores. During the training phase, several crops are sampled from the SR images and their downsampled versions. Features from both scales are extracted via the encoder trained in the pretext phase and concatenated to form a multiscale representation. This fused feature set, along with the ground-truth scores, serves as the input for training the downstream network. At inference time, as illustrated in Fig. \ref{fig:s3riqa_downstream}, we apply the same cropping and feature extraction strategy. The final quality score for each image is obtained by averaging the predicted scores across all crops.

\section{Experimental Results}
\label{sec:results}
\subsection{Practical settings}
The pretext model was trained for 60 epochs using a stochastic gradient descent (SGD) optimizer, featuring a momentum of 0.9 and a weight decay parameter set at $10^{-4}$. A learning rate schedule that employs cosine annealing with warm restarts \cite{loshchilov2016sgdr} was applied, starting from $10^{-3}$. The ConvNeXT-Tiny encoder shown in Fig. \ref{fig:s3riqa} was pretrained, with its weights subsequently fine-tuned. This encoder and its subsequent projection head were configured with channel dimensions of 768 and 128, respectively. The training process incorporated a patch size of $256\times256$, a temperature parameter $\tau$ of 0.1 in equation (\ref{eq:simclr_objective}), and a batch size of 16.

\subsection{Evaluation datasets}
As mentioned above, the pretext training phase uses 15,374 SR images from our SRMORSS dataset for representation learning. The $\mathrm{S^3RIQA}$ model is subsequently evaluated on multiple benchmark datasets (which are different from the data used for pretext  training to realize cross dataset evaluations) that exhibit real-world distortions originating from authentic LR images, rather than synthetically downsampled ones. The datasets mentioned in Section \ref{sec:sriqad} have been used for this purpose. As mentioned there, the RealSRQ dataset \cite{jiang2022single} comprises 60 reference images, each associated with three scaling factors ($\times2$, $\times3$, and $\times4$), reconstructed using 10 SR algorithms, resulting in a total of 1,620 images. Similarly, the SRIJ dataset \cite{beron2020blind} consists of 32 reference images with three scaling factors ($\times2$, $\times3$, and $\times4$), generated by 7 SR methods, which yield 608 images. Furthermore, the SREB dataset \cite{kim2025subjective} includes 30 reference images with two scaling factors ($\times2$ and $\times4$), produced using 7 SR approaches, for a total of 420 images.

\subsection{Evaluation protocol}

To evaluate the performance of the model, Spearman's rank order correlation coefficient (SRCC) and Pearson’s linear correlation coefficient (PLCC) were utilized to measure monotonicity and accuracy, respectively. Each dataset was randomly split into an $80\%$ training set and a $20\%$ test holdout, and the mean performance over 20 iterations is reported. A ridge regressor \cite{hoerl1970ridge} with a regularization parameter $\alpha = 1$ was trained using the ground-truth scores from the training subsets. The regressor received as input the feature representations produced by the encoder, and no fine-tuning of the encoder weights was performed during evaluation. In this stage, image features were extracted at both full and half scales and concatenated to form the final representation, following the strategy proposed in \cite{agnolucci2024arniqa}, which has demonstrated strong effectiveness in prior studies. As proposed in \cite{10203406}, five crops (four corners and the center) were taken at both scales, and the quality scores predicted by the regressor for these crops were averaged to yield the final quality score for each image.

\subsection{Results}

\begin{table}[t!]
	\centering
	\scriptsize	
	\caption{Performance comparison on SRIJ.}
	\begin{tabular}{cccc}
				
		\toprule
				
		Type                    & Method                                 & PLCC           & SRCC           \\

		\midrule
		\multirow{5}{*}{Generic IQA} & 
		PSNR& 0.377 & 0.379 \\
				
		                        & SSIM \cite{wang2004image}               & 0.455          & 0.485          \\
				
		                        & NIQE \cite{mittal2012making}            & 0.556          & 0.491          \\
				
		                        & BRISQUE \cite{mittal2012no}             & 0.727          & 0.716          \\
						
		                        & ARNIQA* \cite{agnolucci2024arniqa}      & 0.935         & \underline{0.926}              \\
				
		\midrule
				
		\multirow{5}{*}{SR-IQA} & $\mathrm{Q_{ODA}}$ \cite{beron2020blind} & 0.896          & 0.863          \\
						
		                        & CN-BSRIQA \cite{rehman2024cn}            & \underline{0.936}          & 0.895          \\
		                        & Zhang \emph{et al}. \cite{zhang2022no} & 0.891          & 0.889          \\
		\cmidrule{2-4}
		                        & $\mathrm{S^3RIQA}$ (Ours)               & \textbf{0.974} & \textbf{0.970} \\
				
		\bottomrule
	\end{tabular}
	\vspace{-4mm}
	\label{tab:srij_results}
\end{table}

\begin{table}[t!]
	\centering
	\scriptsize	
	\caption{Performance comparison on RealSRQ.}
	\begin{tabular}{cccc}
				
		\toprule
				
		Type                         & Method                            & PLCC           & SRCC           \\

		\midrule
		\multirow{7}{*}{Generic IQA} & PSNR                               & 0.081          & 0.095          \\
		                             & SSIM \cite{wang2004image}           & 0.106          & 0.132          \\
		                             & NIQE \cite{mittal2012making}        & 0.027          & 0.098          \\
		                             & BRISQUE \cite{mittal2012no}         & 0.052          & 0.008          \\
		                             & CNNIQA \cite{kang2014convolutional} & 0.671          & 0.667          \\
				
		                             & MANIQA \cite{yang2022maniqa}        & \underline{0.863}          & \textbf{0.781}          \\ 
				
		                             & ARNIQA* \cite{agnolucci2024arniqa}  & 0.759          & 0.569              \\
				
		\midrule
				
		\multirow{5}{*}{SR-IQA}      & NRQM \cite{ma2017learning}          & 0.145          & 0.004          \\
				
		                             & C$^{2}$MT \cite{li2022c}           & 0.718          & 0.704          \\
				
		                             & SGH \cite{fu2023scale}              & 0.685          & \underline{0.707}          \\				
				
		\cmidrule{2-4}
		                             & $\mathrm{S^3RIQA}$ (Ours)          & \textbf{0.913}          & \textbf{0.781}          \\
				
		\bottomrule
	\end{tabular}
	\vspace{-4mm}
	\label{tab:realsrq_results}
\end{table}

\begin{table}[t!]
	\centering
	\scriptsize	
	\caption{Performance comparison on SREB.}
	\begin{tabular}{cccc}
				
		\toprule
				
		Type                    & Method                          & PLCC           & SRCC           \\

		\midrule
		\multirow{4}{*}{Generic IQA} & 
						
		BRISQUE \cite{mittal2012no}         & 0.865             & 0.820             \\
				
		                        & MANIQA \cite{yang2022maniqa}      & 0.844          & 0.808          \\ 
				
		                        & Re-IQA \cite{saha2023re}          & 0.933          & 0.899          \\
		
		                        & ARNIQA \cite{agnolucci2024arniqa} & \underline{0.935} & \underline{0.900} \\
		\midrule
				
		\multirow{4}{*}{SR-IQA} & NRQM \cite{ma2017learning}    & 0.891          & 0.828          \\
				
		                        & KLTSRQA \cite{jiang2022single}           & 0.891          & 0.858          \\
						
		\cmidrule{2-4}
		                        & $\mathrm{S^3RIQA}$ (Ours)        & \textbf{0.942}          & \textbf{0.908}          \\
				
		\bottomrule
	\end{tabular}
	\vspace{-4mm}
	\label{tab:sreb_results}
\end{table}

This section compares the performance of $\mathrm{S^3RIQA}$ against several related NR-IQA methods in terms of PLCC and SRCC on the evaluation datasets mentioned above. The results are reported in Tables \ref{tab:srij_results}, \ref{tab:realsrq_results}, and \ref{tab:sreb_results}. In addition, PSNR and SSIM results are also reported in the tables; although they are FR-IQA metrics, they are widely used as baseline measures in SR–related tasks. However, for the SREB dataset, the computation of PSNR and SSIM is not possible since the corresponding HR reference images are not accessible. The best and second-best values are highlighted in boldface and underlined styles, respectively. Generally, our method outperforms the other generic and SR-IQA methods on all 3 datasets, and even by a large margin for some of them.\footnote{The results for SREB \cite{kim2025subjective}, RealSRQ \cite{lin2024perception}, and SRIJ \cite{zhang2022no, rehman2024cn} were taken from the corresponding cited works.}\par

The experimental results reveal several important insights. The results demonstrate that the proposed SSL approach achieves superior quality prediction performance compared to fully supervised SR-IQA methods, highlighting the effectiveness and generalization capability of the proposed $\mathrm{S^3RIQA}$ framework when trained with the accompanying unsupervised SRMORSS dataset.
The findings clearly establish the strong dependence of perceptual quality on both the chosen SR algorithm and the scaling factor.\par

\begin{table}[t!]
\scriptsize
\centering
\caption{$\mathrm{S^3RIQA}$ performance evaluated on SR results with varying scaling factors}
\begin{tabular}{cccccccc}

\toprule

\multirow{2}{*}{Dataset} & \multicolumn{2}{c}{$\times2$} & \multicolumn{2}{c}{$\times3$} & \multicolumn{2}{c}{$\times4$} \\

& PLCC & SRCC & PLCC & SRCC & PLCC & SRCC \\

\midrule

SREB & 0.909 & 0.836 & - & - & 0.877 & 0.719 \\

RealSRQ & 0.932 & 0.817 & 0.903 & 0.736 & 0.848 & 0.770 \\

SRIJ & 0.920 & 0.907 & 0.900 & 0.899 & 0.912 & 0.901\\

\bottomrule
\end{tabular}
\label{tab:scaling_factor_results}
\end{table}

We further conducted downstream evaluations on subsets of the evaluation datasets, where each subset comprises super-resolution results associated with a specific scaling factor. The results summarized in Table \ref{tab:scaling_factor_results} demonstrate that increasing the scaling factor, corresponding to more challenging SR conditions, generally results in inferior $\mathrm{S^3RIQA}$ performance. Furthermore, the superior PLCC and SRCC observed for the arbitrary scaling factor in the preceding tables compared to these fixed scaling factor experimental results, are consistent with the proposed pretext training strategy, which integrates multiple scaling factors. This design better reflects real-world scenarios, in which the SR scaling factor is typically unknown a priori.

\subsection{Ablation studies}

\begin{table}[t!]
\scriptsize
\centering
\caption{Ablative studies on eliminating auxiliary task and color space transformation}
\begin{tabular}{cccccccc}

\toprule

\multirow{2}{*}{Method} & \multicolumn{2}{c}{SREB} & \multicolumn{2}{c}{SRIJ} & \multicolumn{2}{c}{RealSRQ} \\

& PLCC & SRCC & PLCC & SRCC & PLCC & SRCC \\

\midrule

Base & 0.942 & 0.908 & 0.974 & 0.970 & 0.913 & 0.781 \\

w/o Color Space & 0.927 & 0.889 & 0.968 & 0.963 & 0.914 & 0.780 \\

w/o Auxiliary& 0.937 & 0.898 & 0.976  & 0.972 & 0.912 & 0.786\\

\bottomrule
\end{tabular}
\label{tab:ablation}
\end{table}

To assess the impact of each component, we performed an ablation study for the above mentioned datasets in Table \ref{tab:ablation}. The effectiveness of the auxiliary task and the color space transformation is evaluated across three datasets.
The base model in this table corresponds to Fig. \ref{fig:s3riqa}, where all components are enabled. The impact of removing each component differs across datasets.\par Specifically, on the SREB dataset, both components are beneficial, with the color space transformation having a more pronounced effect than the auxiliary SR scaling factor regression task. On the SRIJ dataset, the color space transformation again improves performance, whereas the auxiliary task provides little benefit; removing it yields results comparable to the base model. Finally, on the RealSRQ dataset, the core proposed SSL strategy alone appears sufficient, and the additional components do not provide further gains. As a practical note, removing the color-space transformation caused the training to fail to converge. This is because of the presence of layer normalizations in both the auxiliary and the projection head, which suppress gradient magnitudes when all inputs lie in the same RGB space. Thus, layer normalizations are disabled in this case.\par

These observations suggest that since SREB and SRIJ differ more substantially from the pretext dataset SRMORSS, the additional components help mitigate the distribution shift. In contrast, RealSRQ, which originates from the same source as SRMORSS (both derived from RealSR), exhibits a smaller distribution shift, and the contrastive learning strategy alone is adequate for learning meaningful pretext representations.

\section{Conclusion}
\label{sec:conclusion}
In this paper, we propose $\mathrm{S^3RIQA}$, the first self-supervised image quality assessment method for super-resolution, along with a new SRMORSS dataset for the pretext training phase. To better reflect real-world super-resolution scenarios, we avoid relying on commonly used synthetically downsampled training images and instead use only genuinely low-resolution data for both training and testing. Moreover, we adopt a no-reference evaluation approach that does not require high-resolution ground-truth images. To construct our dataset, we applied a range of classic and contemporary SR models. Using these data, SSL facilitates learning of unknown SR-specific degradations, independent of the images contents, in an unsupervised manner. It also provides the advantage of requiring only a simple regression model in the final downstream stage.\par

We found that this strategy, together with the proposed deep neural SSL architecture, is highly effective in learning meaningful and discriminative representations of SR degradations in our experimental evaluations on several realistic SR datasets with quality scores, even in the presence of domain shift from pretext training data. An auxiliary network and an additional data transformation process, aimed at improving robustness to SR scaling factors and color space changes, respectively, were incorporated to achieve more accurate results in terms of Pearson and Spearman correlation coefficients. They have shown different effects on the used SR datasets, which is further examined through an ablation study.

A key advantage of the proposed approach over non-SSL methods is its ability to leverage large-scale unlabeled data from diverse applications to customize and enhance the model for specific tasks, without requiring a costly and labor-intensive collection of subjective quality scores. In the future, we plan to apply and adapt $\mathrm{S^3RIQA}$ to real-world SR challenges, such as those found in \cite{anderson2025biology}. Exploring new pretext data transformations and preprocessing strategies to enable more accurate domain-specific SR-distortion manifold learning will be an important next step. We also aim to incorporate geometric cues to support more plausible quality assessment for structured images in specialized application domains. Additionally, integrating more recent SSL techniques, for example methods that more effectively manage negative pairs, may further enhance the proposed model.

\section*{Acknowledgments}
This work has benefited from a French government grant managed by the Agence Nationale de la Recherche under the France 2030 program, reference ANR-23-IACL-0006.
 
%

\bibliographystyle{IEEEtran}
\bibliography{references}

\vfill

\end{document}